\documentclass[sigconf]{acmart}

\usepackage{graphicx}
\usepackage{caption}
\usepackage{mathtools}
\usepackage{color,soul}
\usepackage{algorithm}
\usepackage{algpseudocode}
\usepackage{pifont}
\usepackage{amsfonts}
\usepackage{amsmath}
\usepackage{tabularx, multirow}
\usepackage{bbm}

\usepackage{graphicx}
\usepackage{caption}
\usepackage{mathtools}
\usepackage{color,soul}
\usepackage{algorithm}
\usepackage{algpseudocode}
\usepackage{pifont}
\usepackage{amsfonts}
\usepackage{amsmath}
\usepackage{tabularx}
\usepackage{scalerel,stackengine}
\usepackage{subcaption}

\usepackage{multicol}
\usepackage{graphicx}

\usepackage{float}

\definecolor{Orange}{rgb}{1,0.5,0}
\definecolor{Purple}{rgb}{1,0,1}

\setcopyright{rightsretained}

\acmDOI{10.475/123_4}

\acmISBN{123-4567-24-567/08/06}

\acmConference[WOODSTOCK'97]{ACM Woodstock conference}{July 1997}{El
  Paso, Texas USA} 
\acmYear{1997}
\copyrightyear{2016}

\acmPrice{15.00}

\begin{document}
\title{Efficient Online Inference for Infinite Evolutionary Cluster models with Applications to Latent Social Event Discovery}

\author{Wei Wei}
\affiliation{%
  \institution{Carnegie Mellon University}
}
\email{weiwei@cs.cmu.edu}

\author{Kennth Joseph }
\affiliation{%
  \institution{Northeastern University \footnote{The work was completed while the author was a graduate student at Carnegie Mellon University}
}
}
\email{k.joseph@northeastern.edu}

\author{Kathleen M. Carley}
\affiliation{%
  \institution{Carnegie Mellon University}
}
\email{kathleen.carley@cs.cmu.edu}

%
%
%
%
%
%


\begin{abstract}
The Recurrent Chinese Restaurant Process (RCRP) is a powerful statistical method for modeling evolving clusters in large scale social media data. With the RCRP, one can allow both the number of clusters and the cluster parameters in a model to change over time. However, application of the RCRP has largely been limited due to the non-conjugacy between the cluster evolutionary priors and the Multinomial likelihood. This non-conjugacy makes inference difficult and restricts the scalability of models which use the RCRP, leading to the RCRP being applied only in simple problems, such as those that can be approximated by a single Gaussian emission. In this paper, we provide a novel solution for the non-conjugacy issues for the RCRP and an example of how to leverage our solution for one specific problem - the \emph{social event discovery problem.} By utilizing Sequential Monte Carlo methods in inference, our approach can be massively paralleled and is highly scalable, to the extent it can work on tens of millions of documents. We are able to generate high quality topical and location distributions of the clusters that can be directly interpreted as real social events, and our experimental results suggest that the approaches proposed achieve much better predictive performance than techniques reported in prior work. We also  demonstrate how the techniques we develop can be used in a much more general ways toward similar problems.
\end{abstract}

%
%




\maketitle
\section{Introduction}
With the increasing amount of unlabeled data sets that can be easily acquired, clustering techniques have become increasingly important to the community of machine learning.   Thanks to the growing amount of social media and social networking applications, publicly available text data, specifically, has grown at a massive, exponential rate. As the amount of such data produced has rapidly surpassed human capacity for interpretation of it, one of the most important questions we face today is how we can effectively organize this text into clusters that are meaningful for humans and allow for actionable insights.  

Bayesian ad-mixture based methods such as Latent Dirichlet Allocation (LDA) \cite{blei_latent_2003}, otherwise known as ``topic models'', have been one of the most popular clustering methods for text data. In topic models, latent representations of clusters referred to as ``topics'', which are distributions over words in the corpus, are learned by scanning a large text corpus. Extensions of topic models have been developed for when meta information such as spatial coordinates or timestamps are present. Examples include the geographical topic model\cite{hong2012discovering}, the dynamic topic model \cite{blei2006dynamic} and the event detection model \cite{wei2015bayesian}. In these models, topics usually contain distributions that describe metadata in addition to the word distributions found in traditional topic models.   

When topic models are applied to data over long periods of time, it is likely that topics will change or evolve. For example, a topic about the presidential elections trained on data from 2012 might contain words like ``debate'', ``Democrat'', ``election'', generic to almost any presidential election, and words like ``marriage'' and ``Romney'' that were specific to the 2012 race. Similarly, a topic trained on data from 2016 might contain the same general terms but refer to ``wall'' and ``Trump'' as opposed to ``marriage'' and ``Romney''. A temporal evolutionary model such as the dynamic topic model \cite{blei2006dynamic} can be trained on data from both 2012 and 2016 and can identify a single ``presidential election'' topic that shifted slightly over time between 2012 and 2016, rather than creating two totally distinct topics. Rather than having to perform manual, post-hoc analyses to make such connections, temporal evolutionary topic models thus can easily allow for understanding how general themes of importance in the data shift over time.

Temporal evolutionary topic models can also address the fact that over time, not only do old topics shift, new topics also arise. The non-parametric version of the evolutionary dynamic models uses the Recurrent Chinese Restaurant Process (RCRP)\cite{ahmed2008dynamic} and therefore allows for the number topics to be automatically determined by the data set rather than setting a fixed value. Because the model works on temporally ordered data, inference techniques such as Sequential Monte Carlo can allow for massively paralleled online algorithms to be developed to deal with streaming data sets.  Non-parametric evolutionary dynamic models thus allow for both the creation of ``new'' topics dynamically from the data as well as the evolution of ``old'' topics over time, making them an ideal candidate for the rapid analysis of large-scale text data. 

Unfortunately, such models are difficult to apply to large-scale text data, because inference is too inefficient. One of the primary issues in this inference is the non-conjugacy between the data likelihood and the priors of the evolutionary model. In such models, cluster evolutionary priors are usually chosen to be logistic-normal distributions \cite{blei2006dynamic,ahmed2008dynamic}, which are not conjugate with the Multinomial likelihood used in topic modeling. Non-conjugacy puts significant computational limitations on the evaluation of marginal likelihood, which is usually required for the inference of such statistical models. The usual solution to this issue is to utilize Laplace Approximations to approximate the marginal likelihoods. In this approximation, Taylor expansion up to the second order is used to approximate the integral around a point that maximizes the original function. However, the particular form of the evolutionary dynamic model makes it difficult to solve this maximum point. Based on Bayesian theory, prior work therefore chose the point to maximize the data likelihood  instead of the posterior in order to obtain a tractable solution \cite{ahmed2008dynamic}. This solution, however, ignored information on the prior, which contains historical clusters on previous time steps.   

Another issue with the evolutionary dynamic models for clustering is the difficulties involved in inference in general. Prior work \cite{ahmed2008dynamic} uses RTS smoothing to solve the model, which is only feasible when the emission functions are in the form of strictly Gaussian.  Thus, in situations where emission functions can not be expressed as a single Gaussian, new inferences technique has to be developed.

In this paper, we study inference techniques to solve the evolutionary dynamic clustering problem. To illustrate how our technique works, we apply it onto the \emph{Evolutionary Social Event Discovery (ESED)} problem Based on prior work on event detection\cite{wei2015bayesian}. The ESED task is to discover evolutionary latent clusters of documents that characterize distinct social events by monitoring an evolving set of documents with spatiotemporal meta-data that contain text about social events. Our experimental results suggest that we are able to detect major evolutionary social events on a set of Twitter data. Although the methods are illustrated through a model to solve a specific problem, we note that our inference technique can be used to solve latent evolutionary clustering models in general that are not restricted to the ESED problem.  

\section{Background}

\subsection{Topic Modeling}
Topic modeling has become a popular approach to discover latent topics in large collections of text data\cite{blei_latent_2003,griffiths2004finding}. Over the past decade, a significant amount of work has considered how to extend topic modeling by incorporating metadata\cite{wei2015bayesian}, by improving its sampling efficiencies\cite{li2014reducing,yao2009efficient}, and by improving the generalizability of the model\cite{blei2006dynamic}. Within the topic modeling literature, perhaps the most relevant work for our purposes are those models dealing with temporal dynamics. Specifically, the dynamic topic model \cite{blei2006dynamic} uses a parametric model to characterize changes of topics over time by assuming logistic-normally distributed topics. 

\begin{figure}[!ht]
\includegraphics[width=.5\textwidth]{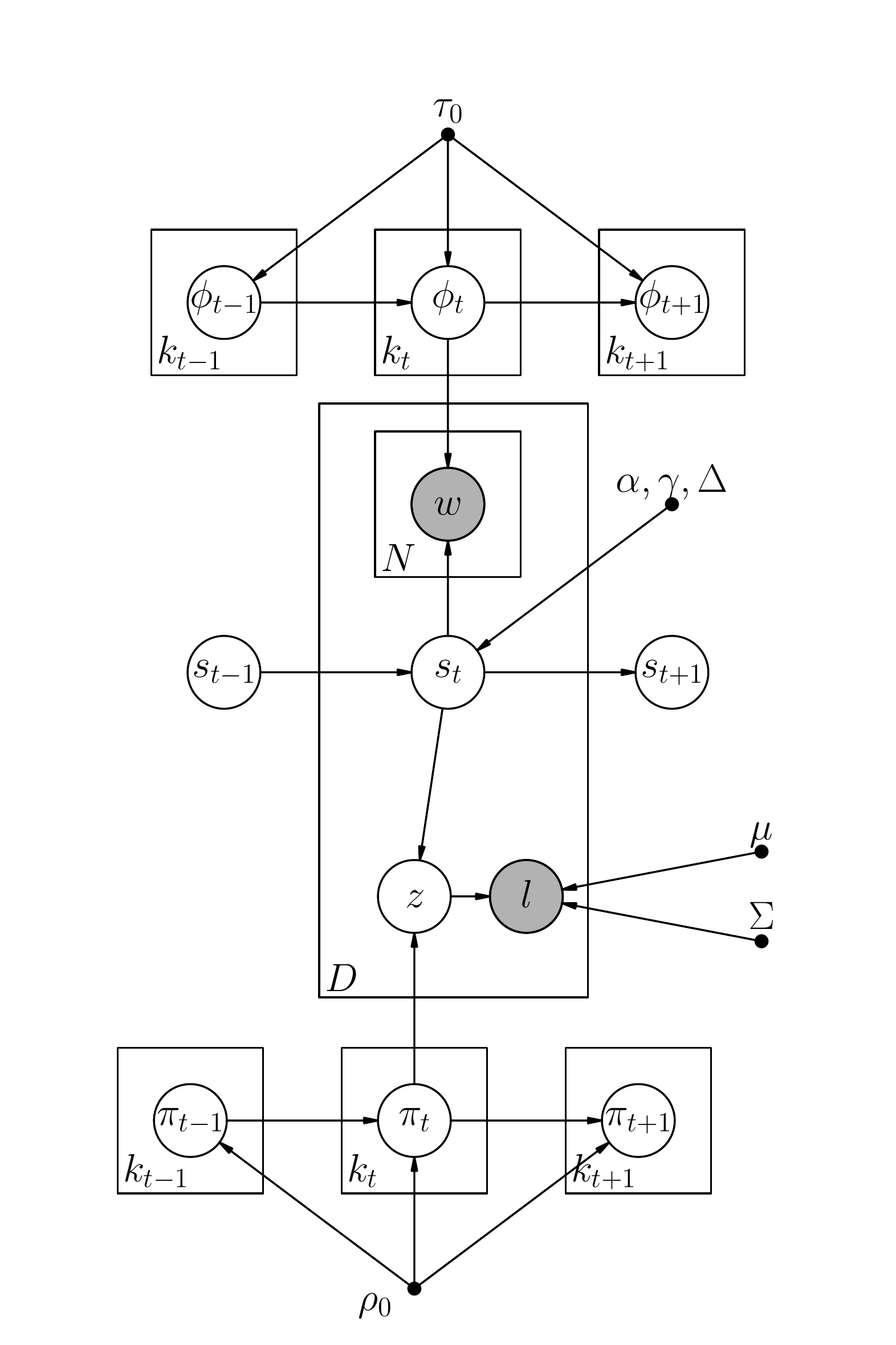}\\
\caption{Graphical Model}
\label{fig2:model}
\end{figure}

\subsection{Non-parametric Bayesian}

There exist a wide range of Bayesian non-parametric techniques that are relevant to topic modeling, most of which are based on Dirichlet Process (DP) \cite{ferguson1973bayesian}. One of the strands of research addresses the temporal dynamics of non-parametric clustering or topic modeling specifically. For example, the recurrent Chinese Restaurant Process (RCRP) \cite{ahmed2008dynamic} divides data into epochs and the process of choosing a specific cluster membership for the $d^{th}$ document at time (epoch) $t$, $s_{t,d}$, is given by Equation~\ref{eq2:def_s}. Here $s_{1:(t,d)-1}$ denotes the set of all documents before (and excluding) the $d^{th}$ document at time $t$. Documents can either create a new cluster with probability proportional to the dispersion parameter $\gamma$ or reuse the existing cluster $k$ with probability proportional to $\sum_{\delta=0}^{\Delta} e^{-\frac{\delta}{\alpha}} m_{t-\delta,k}^{-d}$. Here $m_{t-\delta,k}^{-d}$ is the number of documents belong to cluster $k$ at time $t-\delta$ that includes all the documents before (and excluding) $d$. $e^{-\frac{\delta}{\alpha}}$ here is a decay factor that put more weights on recent time steps rather than historical ones. By using Gaussian transiting distributions, we are able to develop evolutionary document clustering algorithms such as the one in \cite{ahmed2008dynamic}. 

\begin{equation} 
\label{eq2:def_s}
P(s_{t,d} =k| s_{1:(t,d)-1}) \propto 
\begin{cases}
     \sum_{\delta=0}^{\Delta} e^{-\frac{\delta}{\alpha}} m_{t-\delta,k}^{-d} & \text{ k exists }\\
    \gamma,              & \text{k is new}
\end{cases}
\end{equation}

\subsection{Non-conjugacy on Logistic-Normal Prior with Multinomial Likelihood}

Temporal dynamic models with a topic modeling component \cite{ahmed2008dynamic, blei2006dynamic} rely on the logistic normal distribution to provide the ability to model topic evolutions. A logistic normally distributed variable $L(X)$ can be acquired by applying a logistic function $L(\cdot)$ onto the normally distributed variable $X$. Unfortunately, the non-conjugacy between the logistic normal prior and Multinomial likelihood makes it difficult to integrate the topic variable out, which is essential for efficient and effective inference in practice. Many solutions have been proposed to address this issue, such auxiliary sampling using the Polya-Gamma distribution \cite{chen2013scalable} and Laplace approximation \cite{ahmed2008dynamic}. In this paper, we favor the later approach since the auxiliary sampling method still needs to sample each dimension of the latent variable. In the Bayesian setting of the Laplace approximation, our goal is to come up with an approximation to the marginal likelihood, denoted as $M$. The basic idea of Laplace approximation is to use a single point $\widehat{\theta}$ to approximate the whole integral mass. Here we let $h(\theta)=-\frac{1}{N}( \log P(X|\theta) +\log  \pi(\theta) ) $ with $N$ being the number samples, $d$ being the dimension of the data, and $\Sigma = (D^2 h(\widehat{\theta}))^{-1}$.

\begin{equation}	
\begin{aligned} \label{eq2:laplace_classic}
M&=\int P(X|\theta) \pi(\theta) 
\approx 
P(X| \widehat{\theta} ) \pi(\widehat{\theta}) (2\pi)^{d/2} |\Sigma|^{1/2}N^{-d/2}
\end{aligned}
\end{equation}

A Laplace approximation solution that is similar to the problem we are studying in this paper has been proposed in \cite{ahmed2008dynamic}. However, their solution ignored the historical data and, for reasons described below, makes too many simplifying assumptions. We will remedy this issue here by providing a better solution to the approximation that is efficient at the same time.   

\subsection{Sequential Monte Carlo}
Sequential Monte Carlo (SMC) methods, otherwise known as particle filtering \cite{doucet2001introduction} methods, are widely used in the inference of Bayesian models \cite{canini2009online,ahmed2011online,du2015dirichlet}. SMC algorithm keeps track of several sets of instances, known as ``particles'' and update them sequentially. For each instance, an SMC algorithm maintains the posterior distribution of latent variables given the data. In our case, since documents are organized into epochs, SMC maintains the posterior $P(z_{1,(t,d)}, s_{1:(t,d)} | x_{1:(t,d)})$. Here $z_{1,(t,d)}$ is the set of latent variables up to the $d^{th}$ document at time $t$. Similar notations apply to $s_{1:(t,d)}$ and $x_{1:(t,d)}$, which are cluster indicators and the data, respectively. An SMC algorithm updates this posterior to $P(z_{1,(t,d+1)}, s_{1:(t,d+1)} | x_{1:(t,d+1)})$ after scanning another piece of data $x_{(t,d+1)}$ by sampling a proposal distribution in the form of $Q(z_{(t,d+1)}, s_{(t,d+1)} | x_{1:(t,d+1)} , z_{1,(t,d)}, s_{1:(t,d)})$. Here, like in all SMC algorithms, we maintain several sets of those particles and calculate ``particle weights'' to evaluate how good of a representation of the true posterior distribution they are. Once the weights in the particles become unbalanced, we eliminate low particles and duplicate high weight ones. This process is referred to as resampling in the SMC literature \cite{ahmed2008dynamic}. 


\section{Statistical Model}
\begin{table}[t]
\centering
\caption{The notation used in the construction of our statistical model}
\begin{tabularx}{.48\textwidth}{|c|X|} \hline
{\bf Symbol} & {\bf Description}\\ \hline

$(t,d)$ & index of document $d^{th}$ document at time $t$\\ \hline
$1:(t,d)$ & a collection of documents up to the $d^{th}$ document at time $t$\\ \hline
$K_t$ & num. of events at time $t$\\ \hline
$D_t$ & num. of documents at time $t$\\ \hline
$N_{t,d}$ & num. of words belongs to document $(t,d)$\\ \hline
$M$ & num. of Gaussian distributed location centers\\ \hline
$F$ & num. of particles in Sequential Monte Carlo\\ \hline

$s_{t,d}$ & event index of document $(t,d)$\\ \hline
$\pi_{t,k}$ & mixture weight (before logistic transform) of location centers of event $k$ at time $t$\\ \hline
$\phi_{t,k}$ & topic distribution (before logistic transform) of event $k$ at time $t$\\ \hline
$\mu_m$ & mean parameter for location $m$\\ \hline
$\Sigma_m$ & co-variance matrix of component $m$\\ \hline
$l_{t,d}$ & location of document $(t,d)$\\ \hline
$w_{t,d}$ & text that belongs to document $(t,d)$\\ \hline

$\alpha$ & decay factor for RCRP\\ \hline
$\gamma$ & dispersion parameter for RCRP\\ \hline
$\Delta$ & temporal width for RCRP\\ \hline
$\tau_0$ & parameter for topic transition Gaussian co-variance matrix \\ \hline
$\rho_0$ & parameter for location weight Gaussian co-variance matrix\\ \hline

$L(\cdot)$ & logistic function\\ \hline
$\tau_k$ & the first time step when cluster $k$ presents. \\
\hline\end{tabularx}
\label{fig2:notation}
\end{table}

Rooted in prior work on event discovery \cite{wei2015bayesian}, our model characterizes a social event as a collection of distributions on text and location that change with time. Figure~\ref{fig2:model} displays a probabilistic graphical model representation of our model and Table~\ref{fig2:notation} provides an overview of notation used. Our model can roughly be characterized as follows: we assume that a cluster at a particular time step is characterized by a spatial distribution and a topical distribution over words. Importantly, these distributions are allowed to evolve over time. Within a given time step, each document is characterized by the cluster it belongs to. The cluster to which it belongs informs the set of words the document is likely to have, as well as the location the document is likely to be sent from. On the latter point, each document is characterized by a location represented by a latitude, longitude pair. In our model, this latitude and longitude is generated by selecting a specific pre-defined \emph{region}, described below.

A key component of the model we develop is that we discretize the time stamps of tweets (referred to generically here as documents) and organize them into $epochs$. For example, if we chose to discretize our data into month-long time periods, all documents with a timestamp in January, 2016 would fall into the same epoch, while February 2016 will be another epoch, etc. The $d^{th}$ document at time step (or synonymously, epoch) $t$ is labeled with the subscript $(t,d)$. More specifically, each document has a unique \emph{event index} $s_{t,d}$ generated from a RCRP with dispersion parameter $\gamma$, temporal width $\Delta$ and decay factor $\alpha$ \cite{ahmed2008dynamic}. Here $m_{t,k}=\sum_{i=1}^{D_t} \mathbbm{I}(s_{t,i}=k) $ represents the number of documents that belong to cluster $k$ at time $t$ and $m_{t,k}^{-d}$ represents this same quantity up to document $d$. 

Compared to Dirichlet Process \cite{ferguson1973bayesian}, the RCRP considers the temporal dynamics of clusters in the history. Specifically, the hyper-parameter $\Delta$ controls the amount of history information to be taken into account. From Equation~\ref{eq2:def_s}, we can see that recent data will receive a much higher weight and that this weight decays exponentially over time. The parameter $\alpha$ controls the speed of such decay. As a result of RCRP, new events can be ``born'' and old events can ``die out'' once the weight becomes zero, as the event will therefore not be able to attract subsequent documents. 

Within each cluster $k$ in our model, there exists a topical component $\phi_{t,l}$ and a spatial component $\pi_{t,l}$ for each time $t$ that are initially generated by a Gaussian centered on $0$ with diagonal covariance $\tau_0 I$ and $\rho_0 I$ respectively. 

\begin{equation} \label{eq2:phi0}
\phi_{t,k} \sim \mathcal{N} (0,\tau_0 I)
\end{equation}

\begin{equation} \label{eq2:pi0}
\pi_{t,k} \sim \mathcal{N} (0,\rho_0 I)
\end{equation}

For a given document, the probability of generating the words in the document, $w_{(t,d),i}$ and the region index of the document $z_{t,d}$ are determined using a multinomial distribution. By applying a logistic function $L(*)$, the parameters $\phi_{t,l}$ and $\pi_{t,l}$ serve as the natural parameter of these distributions. Hence, $w$ and $z$ follows a logistic normal distribution. Such structure is not new to the community of topic modeling and has been explored by many prior work such as the Correlated Topic Model  \cite{blei_correlated_2007}. 

\begin{equation} \label{eq2:w}
w_{(t,d),i} \sim Multi(L(\phi_{t,s_{t,d}}))
\end{equation}

\begin{equation} \label{eq2:z}
z_{t,d} \sim Multi(L(\pi_{t,s_{t,d}}))
\end{equation}

Once the region index $z_{t,d}$ of a document is determined, the actual document location $l_{t,d}$, which contains a two-dimensional vector representing latitude and longitude, can be generated by using the Gaussian prior $\mu$ and $\Sigma$ generated from each region. 

\begin{equation} \label{eq2:location}
l_{(t,d)} \sim   \mathcal{N} (\mu_z,\Sigma_z)
\end{equation}

One unique characteristic of our model is to allow both the topical parameter $\phi_{t,k}$ and the spatial parameter $\pi_{t,k}$ to evolve with time. This can be achieved by using another Gaussian evolutionary prior on existing events for the current time step that is centered on but that can deviate from the value of the last time step. This idea has been explored in \cite{ahmed2008dynamic}. However, as we mentioned, the authors tried to approximate the emission function using a single Gaussian, which is a reasonable assumption in that model but no longer holds in our scenario since we are modeling spatial component as well. 

\begin{equation} \label{eq2:phit}
\phi_{t,k} \sim \mathcal{N} (\phi_{t-1,k},\tau_0 I)
\end{equation}

\begin{equation} \label{eq2:pi1}
\pi_{t,k} \sim \mathcal{N} (\pi_{t-1,k},\rho_0 I)
\end{equation}

The model can be summarized with a description of its generative process, which is as follows:

\begin{enumerate}
  \item For each time period $t$:
    \begin{enumerate}
    \item For each existing event $k$
    	\begin{enumerate}
    		\item Draw $\pi_{t,k} \sim  \mathcal{N} (\pi_{t-1,k},\rho_0 I)$
    		\item Draw $\phi_{t,k} \sim \mathcal{N} (\phi_{t-1,k},\tau_0 I)$
		\end{enumerate}

    \item For each document $d$
	    \begin{enumerate}

    	 \item Draw event index $s_{t,d}$ from $RCRP (\gamma, \alpha,\Delta)$
    	 \item If $s_{t,d}=k$ is a new event
    	  	\begin{enumerate} 
    		\item Draw $\pi_{t,k} \sim  \mathcal{N} (0,\rho_0 I)$
    		\item Draw $\phi_{t,k} \sim \mathcal{N} (0,\tau_0 I)$
    	 	\end{enumerate}
    	 \item Draw $w_{(t,d),i}$ , $z{t,d}$ and  $l_{t,d}$ according to Eq.\ref{eq2:w}, Eq.\ref{eq2:z}, and Eq.\ref{eq2:location}  
    	 \end{enumerate}
  \end{enumerate}
\end{enumerate}

\section{Scalable Inference}

\subsection{Integrating Variables}
We start with the joint probability of the model and seek a collapsed version of it, $P(s,z,w,l | \mu, \sigma, \gamma, \Delta, \alpha, \rho_0, \tau_0)$ by integrating out the natural parameters $\phi_{t,k}$ and $\pi_{t,k}$. In the following derivations, we will omit hyper-parameters and use ``$\cdot$'' to annotate them for cleaner notation. We also define $g(\cdot)$ to be the likelihood function. The location likelihood $g(\pi_{t,k})$ and the text likelihood  $g(\phi_{t,k})$  are defined in Equation~\ref{eq2:g_pi} and Equation~\ref{eq2:g_phi}, respectively. Here $n^{\pi}_{t,k,g}$ and $n^{\phi}_{t,k,i}$ are the number of occurrence in cluster $k$ at time $t$ for location component $g$ and vocabulary $i$, respectively. 

\begin{equation}
\begin{aligned} \label{eq2:g_pi}
g(\pi_{t,k})&=\prod_{d=1}^{D_t}P(z_{t,d}| \pi_{t,k},s_{t,d}=k)
=
\prod_g
\bigl(
\frac{e^{\pi_{t,k,g}}}
{\sum_j e^{\pi_{t,k,j}}} 
\bigr)
^{n^{\pi}_{t,k,g}}
\end{aligned}
\end{equation}

\begin{equation}
\begin{aligned} \label{eq2:g_phi}
g(\phi_{t,k})&=\prod_{d=1}^{D_t}P(w_{t,d}| \phi_{t,k},s_{t,d}=k)
=
\prod_i
\bigl(
\frac{e^{\phi_{t,k,i}}}
{\sum_j e^{\phi_{t,k,j}}} 
\bigr)
^{n^{\phi}_{t,k,i}}
\end{aligned}
\end{equation}

By utilizing the notations defined above, the integration can be expressed in Equation~\ref{eq2:integration}. Here we use $\tau_k$ to denote the fist time step when cluster $k$ occurs. We also define $\psi^{\pi}_{t,k}=0$ when $t=\tau_k$ and $\psi^{\pi}_{t,k}=\pi_{t-1,k}$ if $t>\tau_k$. A similar definition is applied to $\psi^{\phi}_{t,k}$. 

\begin{equation}
\begin{aligned} \label{eq2:integration}
P&(s,z,w,l | \cdot)\\
=&\prod_{k=1}^K \prod_{t=\tau_k}^T \int_{\pi_{t,k}} 
g(\pi_{t,k}) P(\pi_{t,k} |\psi^{\pi}_{t,k})
 \int_{\phi_{t,k}} 
g(\phi_{t,k}) P(\phi_{t,k} |\psi^{\phi}_{t,k})\\
&\prod_{t=1}^T\prod_{t=\tau_k}^T \prod_{d=1}^{D_t} P(s_{t,d} | s_{1:(t,d)-1})
P(l_{t,d}| \mu_k, \Sigma_k,z_{t,d}=k)
\end{aligned}
\end{equation}

The key to the integration is to correctly deal with terms involving the likelihood function $g(\cdot)$ and its priors,  $\prod_{t=\tau_k}^T \int_{\pi_{t,k}} 
g(\pi_{t,k}) P(\pi_{t,k} |\psi^{\pi}_{t,k})$. As we will see shortly, we can conduct integrations in a chain fashion from the very beginning when $t=\tau_k$ all the way to the end when $t=T$. When an integration is done for a specific time step, we will get a constant term and a ``future term''. The future term, which we denote as $f_{t,k}(\pi_{t+1,k} | \theta_{t,k})$ contains the information for a future integration and will participate in the integration for the next time step. The constant term, which we annotate as $D_{t,k}$ will be emitted  as part of our final integration result. Here we will focus on the terms that involves $\pi$ and we will omit the procedures for $\phi$ since it can be derived similarly.

As we mentioned above, the future term $f_{t,k}(\pi_{t+1,k} | \theta_{t,k})$  is generated as part of the integration result at time $t$. It contains variable $\pi_{t+1,k}$ that will participate the integration of the next time step $t+1$ with parameter $\theta_{t,k}$ that is determined by information on the previous time steps. To illustrate how the future term $f_{t,k}(\pi_{t+1,k} | \theta_{t,k})$ interplay with the integration, we define the following relationship in Equation~\ref{eq2:future_term_relation}. We also assume that $f_{t-1,k}(\pi_{t,k} | \theta_{t-1,k})$ is in the form of Gaussian distribution with mean $\theta_{t-1,k}$ and covariance matrix $\rho_0 I$. We will prove this using mathematical induction. 

\begin{equation}	
\begin{aligned} \label{eq2:future_term_relation}
\int_{\pi_{t,k}} 
g(\pi_{t,k}) P(\pi_{t+1,k} |\pi_{t,k}, \rho_0 I) f_{t-1,k}(\pi_{t,k} | \theta_{t-1,k})
\end{aligned}
\end{equation}

For the base cases, where $t=\tau_k-1$ we define $f_{\tau_k-1,k}(\pi_{t=\tau_k,k} | \theta_{\tau_k-1,k})$ to be a zero mean Gaussian with covariance matrix $\rho_0 I$. One can validate this definition by taking $f_{\tau_k-1,k}(\pi_{t=\tau_k,k} | \theta_{\tau_k-1,k})$ into Equation~\ref{eq2:future_term_relation} to get the expression for the first integration.   

\begin{equation}	
\begin{aligned} \label{eq2:base_case_f}
f_{\tau_k-1,k}(\pi_{t=\tau_k,k} | \theta_{\tau_k-1,k})=  \mathcal{N} (\pi_{t,k} | 0, \rho_0 I)
\end{aligned}
\end{equation}

For the general case where $\tau_k\leq t < T$, we define the recursive formula of $f_{t,k} (\cdot)$ in Equation~\ref{eq2:recursive_f} to be the integration of $\pi_{t,k}$ divided by a constant $D_{t,k}$, which is defined in Equation~\ref{eq2:definition_d} and is designed to absorb all constants that are not related to the Gaussian distribution to ``participate'' in the next round of integration. 

\begin{equation}	
\begin{aligned} \label{eq2:definition_d}
D_{t,k}=
\mathcal{N}(\widehat{\pi_{t,k}} | \theta_{t-1,k}, \rho_0 I)(2\pi)^{d/2}|\Sigma_{t,k}|^{1/2}N_{t,k}^{-d/2}g(\widehat{\pi_{t,k}})
\end{aligned}
\end{equation}

Here we utilize the induction assumption that $f_{t-1,k}$ is a Gaussian distribution with mean $\theta_{t-1,k}$ and covariance matrix $\rho_0 I$. We also use Laplace Approximation to approximate the integral around a point $\widehat{\pi_{t,k}}$, which will be discussed in more detail in the next sub-section. After letting $D_{t,k}$ to absorb all the constants, we again get a Gaussian form of $f_{t,k} (\cdot)$ with mean value equal to $\widehat{\pi_{t,k}}$ and covariance matrix $\rho_0 I$. 
\begin{equation}	
\begin{aligned} \label{eq2:recursive_f}
f_{t,k}&(\pi_{t,k} | \theta_{t-1,k})\\
&= \frac{\int_{\pi_{t,k}} P(\pi_{t+1,k}|\pi_{t,k}, \rho_0 I)\cdot 
f_{t-1,k} ( \pi_{t,k} | \theta_{t-1,k}) g(\pi_{t,k})}{ D_{t,k}}\\
&=\frac{\int_{\pi_{t,k}} \mathcal{N}(\pi_{t,k} | \frac{\pi_{t+1,k}+\theta}{2}, \rho_0 I/2)
\mathcal{N}(\pi_{t+1,k | \theta_{t-1,k}, 2\rho_0I})
g(\pi_{t,k})}{ D_{t,k}}\\
&=\mathcal{N}(\widehat{\pi_{t,k}} | \frac{\pi_{t+1,k}+\theta_{t-1,k}}{2}, \rho_0 I/2)\mathcal{N}(\pi_{t+1,k | \theta_{t-1,k}, 2\rho_0I})\\
&\hspace{0.42cm}\frac{g(\widehat{\pi_{t,k}}) (2\pi)^{d/2}|\Sigma|^{1/2}N^{-d/2}
}{D_{t,k}}\\
&=\mathcal{N}(\pi_{t+1,k} | \widehat{\pi_{t,k}}, \rho_0 I) \mathcal{N}(\widehat{\pi_{t,k}} | \theta_{t-1,k}, \rho_0 I)\\
&
\hspace{0.42cm}\frac{
(2\pi)^{d/2}|\Sigma_{t,k}|^{1/2}N_{t,k}^{-d/2}g(\widehat{\pi_{t,k}})
}
{D_{t,k}}\\
&=\mathcal{N}(\pi_{t+1,k} | \widehat{\pi_{t,k}}, \rho_0 I)
\end{aligned}
\end{equation}

Note that we do not define $f_{t,k} (\cdot)$ when $t=T$ since there will be no term to contribute to future integrations. To summarize, the integration of $\pi_{t,k}$ over time equals to the $\prod_{t=\tau_k}^{T} D_{t,k}$. We can use the same technique to get the integration of $\phi_{t,k}$ to be $\prod_{\tau_k}^T C_{t,k}$ with $C_{t,k}$ defined below:

\begin{equation}	
\begin{aligned} \label{eq2:definition_c}
C_{t,k}=
\mathcal{N}(\widehat{\phi_{t,k}} | \theta_{t,k}, \rho_0 I)(2\pi)^{d/2}|\Sigma|^{1/2}N^{-d/2}g(\widehat{\phi_{t,k}})
\end{aligned}
\end{equation}

We then use the same notation to get the joint distribution after $\pi$ and $\phi$ are integrated out by taking the results in the previous steps into Equation~\ref{eq2:integration}:
\begin{equation}	
\begin{aligned} \label{eq2:joint_distribution_finished}
P(s,z,w,l|\cdot)=&\prod_{t=1}^{T} \prod_{k=1}^{K_t}C_{t,k} D_{t,k}\\
&\prod_{t=1}^{T}\prod_{d=1}^{D_t}P(s_{t,d} | s_{1:(t,d)-1})
 P(l_{t,d}| \mu,\Sigma, z_{t,d})\\
\end{aligned}
\end{equation}

The collapsed joint distribution thus leaves only two variables to be inferred: $z_{t,d}$ and $s_{t,d}$ for each document $(t,d)$. In experiments we found that the MCMC converges very quickly and only several Gibbs iteration steps are necessary for the algorithm to reach convergence.

\subsection{Laplace Approximation to Marginal Likelihood}

Although we have discussed the general form of the joint distribution after the integration, we haven't covered the details on how we conducted the Laplace Approximation when taking the integral. As seen in Equation~\ref{eq2:laplace_classic}, Laplace's method approximates the integral around a specific point where the majority of the probability mass lies on. In our case, $h(\cdot)$ takes the form of a negative log of a Multinomial likelihood function with a Gaussian prior.

Ideally, we should choose $\widehat{\pi_{t,k}}$ to minimize $h(\pi_{t,k})$. However, when we use sequential techniques to solve the model, we do not have the knowledge of cluster parameters in the next time step, $\pi_{t+1,k}$, which is required to evaluate $h(\pi_{t,k})$. Instead, we use the expectation of its prior information $\pi_{t-1,k}$ to approximate $h(\pi_{t,k})$. $h(\pi_{t,k})$ then becomes: 
\begin{equation}	
\begin{aligned} \label{eq2:h_pi}
h(\pi_{t,k})=&
\frac{-\log
\bigl(
\mathcal{N}(\pi_{t,k} | \frac{\theta_{t-1,k} + \pi_{t+1,k}}{2},\frac{\rho_0^2}{2}I  )
g(\pi_{t,k})
\bigr)
}{N_{t,k}}\\
=&
\frac{-\log
\bigl(
\mathcal{N}(\pi_{t,k} | \theta_{t-1,k} ,\frac{\rho_0^2}{2}I  )
g(\pi_{t,k})
\bigr)
}{N_{t,k}}
\end{aligned}
\end{equation}

When the sample size $N_{t,k}$ is large enough, the impact of the prior will be very small and a natural selection of $\widehat{\pi_{t,k}}$ will be the one that maximize its likelihood. This solution is illustrated in Equation~\ref{eq2:solution1} as \textbf{Solution 1} and is used by \cite{ahmed2008dynamic}. Here, we illustrate the solution by its logistic form rather than its original form, which is more useful since $g(\pi_{t,k})$ utilizes the logistic form of $\pi_{t,k}$. This solution simply normalizes the number of documents having the locations in spatial component $i$ for each cluster $k$ at time $t$, $N^{\pi}_{t,k,i}$ with the total number of documents that belong to cluster $k$ at time $t$, $N^{\pi}_{t,k}$. However, this solution ignores all the historical data before time $t$ since it ignored the prior information. 

\begin{equation}	
\begin{aligned} \label{eq2:solution1}
Solution1: 
\frac{e^{\widehat{\pi_{t,k,i}}}}
{\sum_j e^{\widehat{\pi_{t,k,j}}}}
=\frac{N^{\pi}_{t,k,i}}{N^{\pi}_{t,k}}
\end{aligned}
\end{equation}

Another solution that can be used as an natural comparison to Solution 1 is to use the document count of all historical data that cluster $k$ has on this location component $i$ instead of the count just on this particular time step. Equation~\ref{eq2:solution2} illustrates the exact form of \textbf{Solution 2}. Here the solution is taken to be the normalized count of all the documents belong to cluster $k$ that are located in location component $i$, $N^{\pi}_{k,i}$. This solution, however, ignores the temporal importance and information across all time steps are treated equally. 

\begin{equation}	
\begin{aligned} \label{eq2:solution2}
Solution2: 
\frac{e^{\widehat{\pi_{t,k,i}}}}
{\sum_j e^{\widehat{\pi_{t,k,j}}}}
=\frac{N^{\pi}_{k,i}}{N^{\pi}_{k}}
\end{aligned}
\end{equation}

However, we note that neither of the solutions above take into account of the prior information. A better approach is to solve $\widehat{\pi_{t,k}}$ to minimize the whole $h(\pi_{t,k})$ rather than only the likelihood part. In order to do this, we take the derivative of Equation~\ref{eq2:h_pi} and set it to zero. After we assume that $\sum_j e^{\pi_{t,k,j}}=1$, we are left with the relations in Equation~\ref{eq2:pi_optimal_solving}.

\begin{equation}	
\begin{aligned} \label{eq2:pi_optimal_solving}
\frac{2\theta_{t,k,i} + n_{t,k,i}\rho_0^2}
{n_{t,k}}
-\frac{2}
{\rho_0^2n_{t,k}} \pi_{t,k,i}
=e^{\pi_{t,k,i}}
\end{aligned}
\end{equation}

A novel observation of the present work is that the above equation falls into the set of problems that can be solved using the notation of \emph{Lambert's W} \cite{corless1996lambertw}. Conveniently, this solution can be expressed analytically and we illustrate it in Equation~\ref{eq2:lambert} in the logistic form. In this equation, we define $n^{'}_{t,k,i}$ such that $e^{\theta_{t,k,i}} = \frac{n^{'}_{t,k,i}}{n_{t,k}}$ to represent the pseudo counting that is introduced by the prior, an important trick that will be utilized later.
 
\begin{equation}	
\begin{aligned} \label{eq2:lambert}
\frac{e^{\widehat{\pi_{t,k,i}}}}
{\sum_j e^{\widehat{\pi_{t,k,j}}}}
&=
\frac{2}{\rho_0^2 n_{t,k}}
W(
\frac{
e^{\theta_i} \rho_0^2 n_{t,k}
}
{
2
}
e^{\frac{ \rho_0^2 n_{t,k,i} }{2}}
)\\
&=
\frac{2}{\rho_0^2 n_{t,k}}
W(
\frac{
\rho_0^2 n^{'}_{t,k,i}
}
{
2
}
e^{\frac{\rho_0^2 n_{t,k,i} }{2}}
)
\end{aligned}
\end{equation}

We observed that terms inside Lambert W's function can be bounded by two quantities.

\begin{equation}	
\begin{aligned} \label{eq2:lambert_relation}
&\min\{
\frac{
\rho_0^2 n_{t,k,i}
}
{
2
}
e^{\frac{ \rho_0^2n_{t,k,i} }{2}}
,
\frac{
\rho_0^2 n^{'}_{t,k,i}
}
{
2
}
e^{\frac{ \rho_0^2 n^{'}_{t,k,i} }{2}}
\}\\
&\leq
\frac{e^{\widehat{\pi_{t,k,i}}}}
{\sum_j e^{\widehat{\pi_{t,k,j}}}}
\leq\\
&\max\{
\frac{
\rho_0^2 n_{t,k,i}
}
{
2
}
e^{\frac{ \rho_0^2 n_{t,k,i} }{2}}
,
\frac{
\rho_0^2 n^{'}_{t,k,i}
}
{
2
}
e^{\frac{ \rho_0^2 n^{'}_{t,k,i} }{2}}
\}\\
\end{aligned}
\end{equation}

By utilizing the fact that $W(xe^{x})=x$, we know that the actual solution of 
$\frac{e^{\widehat{\pi_{t,k,i}}}} {\sum_j e^{\widehat{\pi_{t,k,j}}}}$ must lie in the linear combination of its lower and upper bounds. A good choice of the linear weight is the use the information on the variance, $\rho_0$. Since $\rho_0$ controls the amount of information that we can allow to change from one time step to the other, a natural choice of the combination weight would be $1/(1+\rho_0)$ and $\rho_0/(1+\rho_0)$. We put weight $1/(1+\rho_0)$ on the bound that contains the information about the prior while using $\rho_0/(1+\rho_0)$ on the bound that contains the information on the current time step. Our solution, which here will be referred to as \textbf{Solution 3}, takes the linear combination of the document count $n_{t,k,i}$ at time $t$ of cluster $k$ on component $i$ and the pseudo document count on the prior $n^{'}_{t,k,i}$ normalized by the total number of documents that belong to cluster $k$ on this time step $t$, $n_{t,k}$. And since $n^{'}_{t,k,i}$ is normalized by $n_{t,k}$, our solution takes value range from 0 to 1. 

\begin{equation}	
\begin{aligned} \label{eq2:lambert_final}
& Solution 3: \frac{e^{\widehat{\pi_{t,k,i}}}}
{\sum_j e^{\widehat{\pi_{t,k,j}}}}
\\
&=
\frac{2}
{\rho_0^2 n_{t,k}}
\bigl(
\frac{1}{1+\rho_0}
W(
\frac{
\rho_0^2 n_{t,k,i}
}
{
2
}
e^{\frac{n_{t,k,i} \rho_0^2}{2}}
\bigr)
+
\frac{\rho_0}{1+\rho_0}
W(
\frac{
\rho_0^2 n^{'}_{t,k,i}
}
{
2
}
e^{\frac{n^{'}_{t,k,i} \rho_0^2}{2}}
)
)\\
&=
\frac{2}
{\rho_0^2 n_{t,k}}
(
\frac{
1
}
{1+\rho_0}
\frac{
\rho_0^2 n_{t,k,i}
}
{
2
}
+
\frac{
\rho_0
}{1+\rho_0}
\frac{
\rho_0^2 n^{'}_{t,k,i}
}
{
2
}
)\\
&=
\frac{
\frac{
1
}
{1+\rho_0}
n^{'}_{t,k,i}
+
\frac{
\rho_0
}
{1+\rho_0}
 n_{t,k,i}
}
{n_{t,k}}
\end{aligned}
\end{equation}
 
We can derive a similar solution for the optimal points to be used in Laplace approximation for $\phi_{t,k}$. For the length of this paper we will omit the exact derivation since the results are highly similar.

\subsection{Sample Cluster Index $s_{t,d}$}

Starting from Equation~\ref{eq2:joint_distribution_finished}, it is now straightforward to derive Equation~\ref{eq2:sample_s} to sample the cluster index $s_{t,d}$ for each document. Here we see that the equation is linear to the number of words in the document, $N_{t,d}$. We need to choose one of the three solutions proposed in the previous section to substitute $\widehat {\phi_{t,k,n}}$ and $\widehat {\pi_{t,k,z}}$. $P(s_{t,d} | s_{1:(t,d)-1})$ is the RCRP prior defined in Equation~\ref{eq2:def_s}.
\begin{equation}	
\begin{aligned} \label{eq2:sample_s}
P&(s_{t,d}=k | s_{1:(t,d)-1} , w_{t,d}, z_{t,d})\\
 & \propto
 \prod_{n=1}^{N_{t,d}}  
 \bigl(
 \frac{e^{\widehat {\phi_{t,k,w_{t,d,n}}}}}
 {\sum_j e^{\widehat {\phi_{t,k,j}}}}
 \bigr) ^{N_{d,t,i}}
 \bigl(
 \frac{e^{\widehat {\pi_{t,k,z_{t,d}}}}}
 {\sum_j e^{\widehat {\pi_{t,k,j}}}}
 \bigr)
 P(s_{t,d} | s_{1:(t,d)-1})
\end{aligned}
\end{equation}

\subsection{Sample Region Index $z_{t,d}$}
Similarly, we can derive the equation to sample the location region index $z_{t,d}$ for each document from the joint distribution in Equation~\ref{eq2:joint_distribution_finished}. Here we see that the probability of selecting $z_{t,d}$ is proportional to the logistic normal component $ \frac{e^{\widehat {\pi_{t,k,z}}}}
 {\sum_j e^{\widehat {\pi_{t,k,j}}}}$
and $P(l_{t,d} | \mu_z, \Sigma_z)$, which is the Gaussian probability of location $sl_{t,d}$ on the $z_th$ component of the Gaussian prior.

\begin{equation}	
\begin{aligned} \label{eq2:sample_z}
P(z_{t,d} =z | s_{t,d}=k, l_{t,d})
\propto
 \bigl(
 \frac{e^{\widehat {\pi_{t,k,z}}}}
 {\sum_j e^{\widehat {\pi_{t,k,j}}}}
 \bigr)
 P(l_{t,d} | \mu_z, \Sigma_z)
\end{aligned}
\end{equation}

\subsection{SMC Updates}
As we stated in the background section, SMC evaluates a weight for each particle and we need to update this weight every time after we have sampled a new document. Our result is illustrated in Equation~\ref{eq2:update_weight}. Here we see that the weight update is proportional to the likelihood of the newly sampled data. 

\begin{equation}	
\begin{aligned} \label{eq2:update_weight}
\omega_{1:(t,d)}^f \propto & 
\omega_{1:(t,d)-1}^f 
\prod_{n=1}^{N_{t,d}}
P(w_{t,d,n}=v| s_{t,d}=k, \phi_{t,k,v})\\
&P(l_{t,d} | z_{t,d}=z, \mu_z,\Sigma_z) 
\end{aligned}
\end{equation}

\subsection{Algorithm}
The general procedure of our approach is illustrated in Algorithm~\ref{alg3:CHalgorithm}. We organize our data into epochs and for each epoch $t$ we process documents one by one. For each document $(t,d)$, only two variables $z$ and $s$ are sampled and we iterate through the MCMC step $MaxIter$ times. Particle weights are then updated and we evaluate whether it is necessary to resample particles by comparing the L2 norm of the particle weight to a threshold.

\begin{algorithm}
\caption{Particle Filtering Algorithm Framework}
\label{alg3:CHalgorithm}
\begin{algorithmic}[1]
\State Initialize $\omega_1^f$ to $\frac{1}{F}$ for all $f \in \{1,..F\}$
\For{epoch $t$ from $1$ to $T$}
\For{document $d$ from $1$ to $D_t$}
\For{particle $f\in \{1,..F\}$}
\For{iter $\in MaxIter$ }
\State Sample $s$,$z$ using Eq.~\ref{eq2:sample_s} and \ref{eq2:sample_z}
\EndFor
\State Update $\omega^f$ using Eq.~\ref{eq2:update_weight}
\EndFor
\State Normalize particle weight $\omega^f$
\If{$||\omega_t||_2^{-2}< $ threshold}
\State resample particles
\EndIf
\EndFor
\EndFor
\end{algorithmic}
\end{algorithm}

\section{Data}

We collected Twitter data from August, 2010 to September, 2012 using Twitter's Decahose API. Only tweets with geo-coordinates within United States are kept; everything else is discarded. We conducted basic natural language processing on the data and deleted stop words and punctuations. Words in the documents are converted to lower cases and are tokenized. Low frequency words that appear below a threshold are deleted. At the end, we are left with a dictionary size of 40,173 unique tokens and a document size of 5,298,978. We keep 90$\%$ of them for training and $10\%$ of them for testing.    

\begin{table} 
\centering
\caption{Summary of dataset used in the experiment}
\begin{tabular}{|c|c|} \hline
Item& Statistics\\ \hline
Spatial Coverage &  United States\\ \hline
Temporal Coverage &  Aug, 2010 to Sep, 2012\\ \hline
Vocabulary Size &  40,173 \\ \hline
Num. Documents & 5,298,978\\ 
\hline\end{tabular}
\label{tab2:dataset}
\end{table}

\section{Experimental Results}
\subsection{Qualitative Results}

\begin{figure*}
\begin{multicols}{2}
  \begin{subfigure}[b]{\columnwidth}
    \includegraphics[width=\columnwidth]{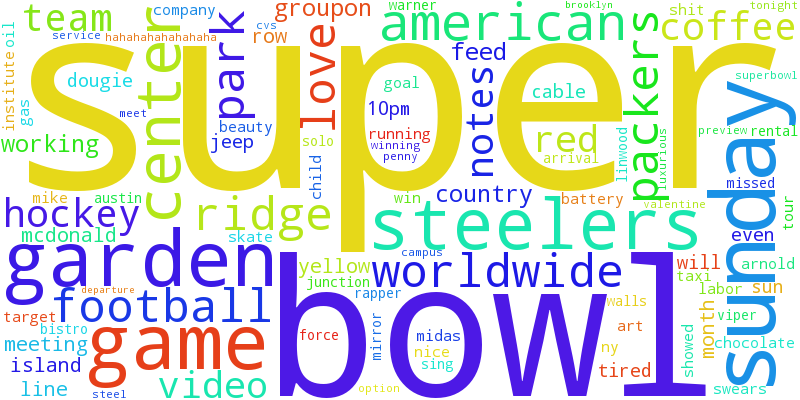}\par 
    \includegraphics[width=\columnwidth]{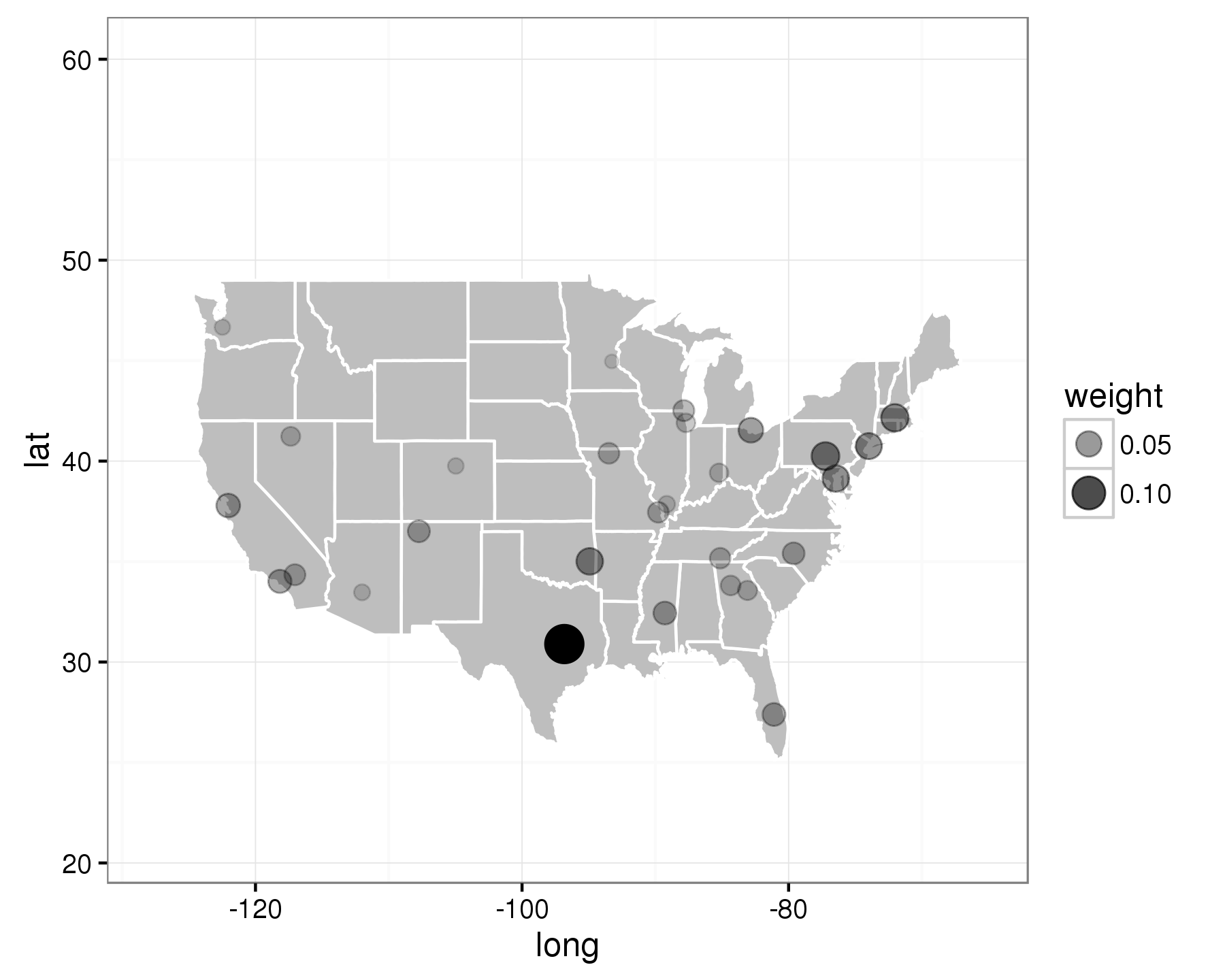}\par 
      \caption{Superbowl event on February 2011}
        \label{fig2:Superbowl 2011}
    \end{subfigure}

  \begin{subfigure}[b]{\columnwidth}
    \includegraphics[width=\columnwidth]{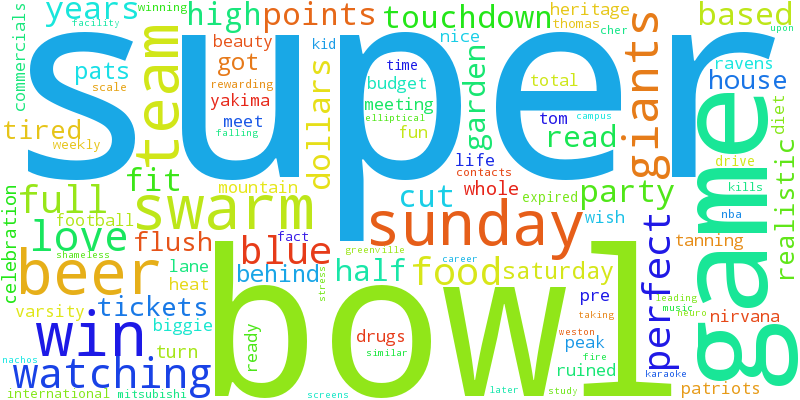}\par 
    \includegraphics[width=\columnwidth]{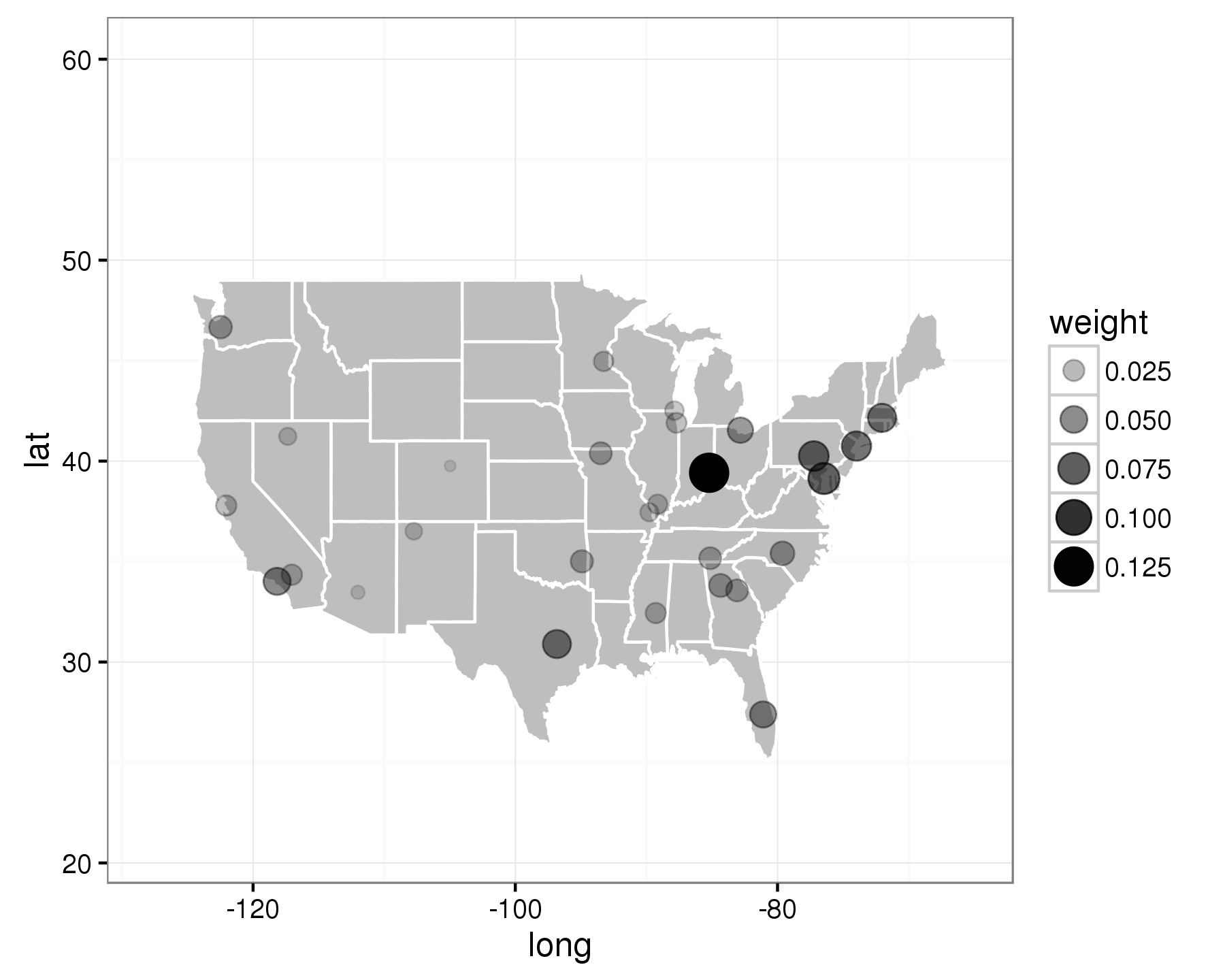}\par 
      \caption{Superbowl event on February 2012}
        \label{fig2:Superbowl 2012}
    \end{subfigure}
\end{multicols}
\caption{Superbowl Event Detected by Our Algorithm}
\label{fig2:qualitative}
\end{figure*}

We first evaluate our model qualitatively by manually examining the contents of various topics (clusters) uncovered by the model during out experiment. Figure~\ref{fig2:qualitative} illustrates one particular cluster that demonstrates the ability of the model to capture an evolving spatio-temporal topic focused on the Super Bowl, the championship game of the National Football League (NFL). As a convenience, we also list a fact sheet about the Super Bowls in these two years, which can be found in Table~\ref{tab2:fact_sheet}. Shown in Figure~\ref{fig2:qualitative} are two pairs of summary figures (split by the left and right) characterizing the parameterizations of a single topic detected by the algorithm we propose. The pair of figures on the left characterizes the representative distributions of this topic in February 2011, the pair on the right characterizes these same distributions in February 2012.  Each pair of figures is characterized by a word cloud (top) that represents the distribution of words for the topic and a map (bottom) with dots representing the spatial distribution of the topic in terms of weights in its mixture distribution over the modeled spatial centers. 

Clearly, the two events characterized in Figure~\ref{fig2:qualitative} share many common terms relevant to the Super Bowl in general - ``super'', ``bowl'', and ``Sunday'' being examples.  However, at these two different points in time, this topic also exhibits important distinct properties.  In the word cloud visualization of the 2011 Super Bowl (top left of Figure~\ref{fig2:qualitative}, we see words such as ``Steelers'' and ``Packers'', which were the names of the two competing teams in that year. Further, the spatial distribution of the event topic at this time clearly highlights an area in central Texas, which was the actual location at which the Super Bowl was held event is held at that year. Since the Super Bowl is a nationwide event, the method also captures minor levels of activity across the country. 

In contrast, on the right hand side of Figure~\ref{fig2:qualitative}, we see terms relevant to the 2012 Super Bowl, including the names of the two teams playing in it (``Giants'' and ``Patriots''). Importantly, teams from the previous year are no longer important to the topic, even though terms relevant to the Super Bowl itself have been retained. Similarly, the spatial visualization clearly illustrated the spatial center of the event, which lies in the state of Indiana, but as the Super Bowl was still a nationwide event in 2001, location of interests exist across the country.  This is particularly true in 2012, relative to 2011, in the New York City and Boston areas, which are where the two teams playing in the game were from.

This analysis of the output of our model gives us confidence that the model can capture events of interest in the manner expected. We now turn to a quantitative analysis of model performance on several different prediction problems.

\begin{table} 
\centering
\caption{Fact sheet about the Superbowl Event}
\begin{tabular}{|c|c|c|} \hline
 & Superbowl XLV&  Superbowl XLVI\\ \hline
Time & February 6, 2011&February 5, 2012\\ \hline
Location &Arlington, TX &  Indianapolis, IN\\ \hline
\multirow{ 2}{*}{Teams}  &Pittsburgh Steelers&New York Giants \\
&Green Bay Packers&New England Patriots  \\ 
\hline\end{tabular}
\label{tab2:fact_sheet}
\end{table}

\subsection{Numerical Results}

We conducted numerical results by first training our model using $90\%$ of the data and then testing it on the rest of the data set by measuring its testing perplexity and the Mean Square Error (MSE) of prediction on held out document locations. We compared three different solutions for Laplace approximation in Eq.~\ref{eq2:solution1}, Eq.~\ref{eq2:solution2} and Eq.~\ref{eq2:lambert_final}. Although our theoretical results favor \textbf{Solution 3}, comparison is still important since prior work reported using \textbf{Solution 1} and \textbf{Solution 2} in similar models. \\

\begin{figure*}
\begin{multicols}{4}
  \begin{subfigure}[b]{0.24\textwidth}
    \includegraphics[width=\columnwidth]{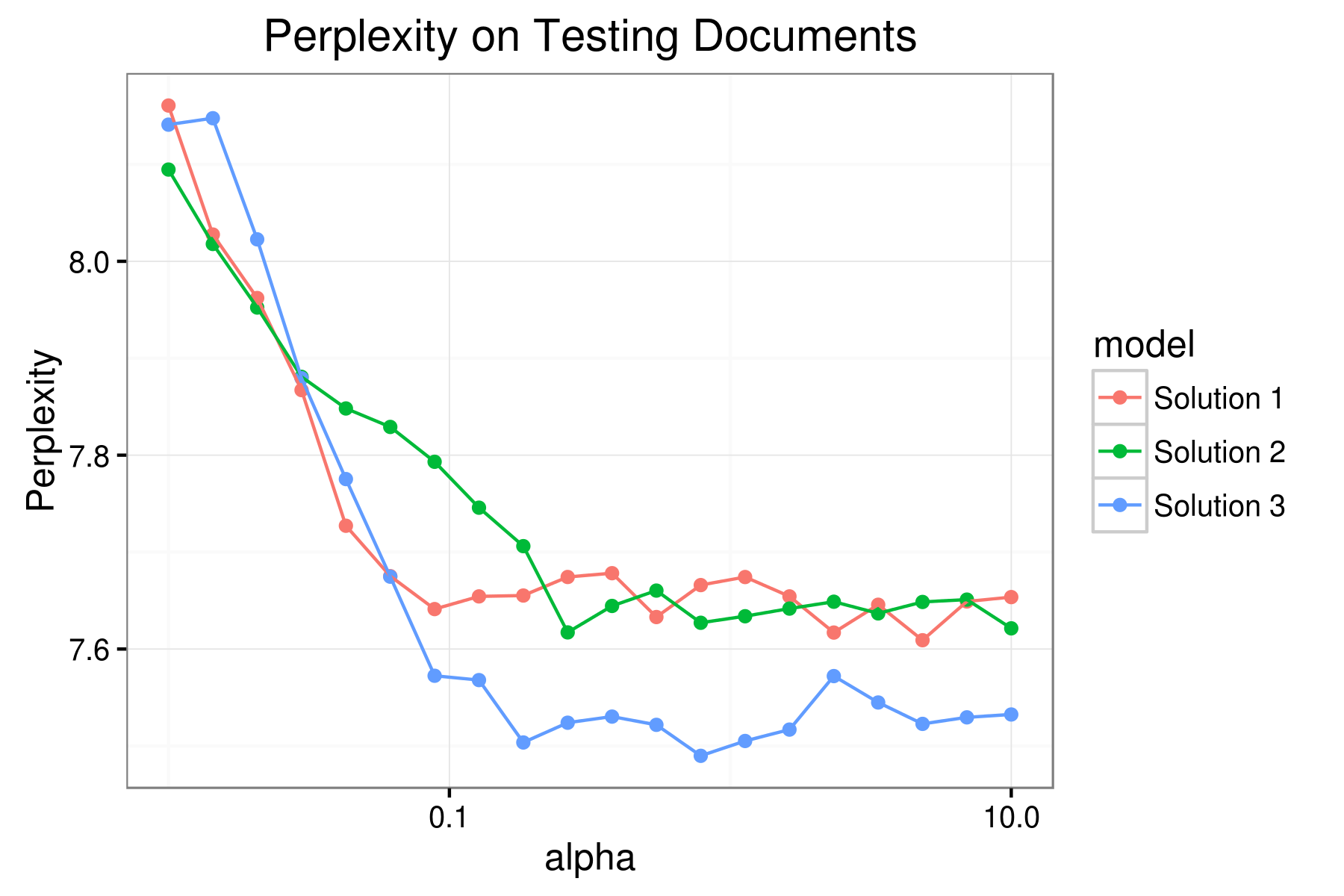}\par 
    \includegraphics[width=\columnwidth]{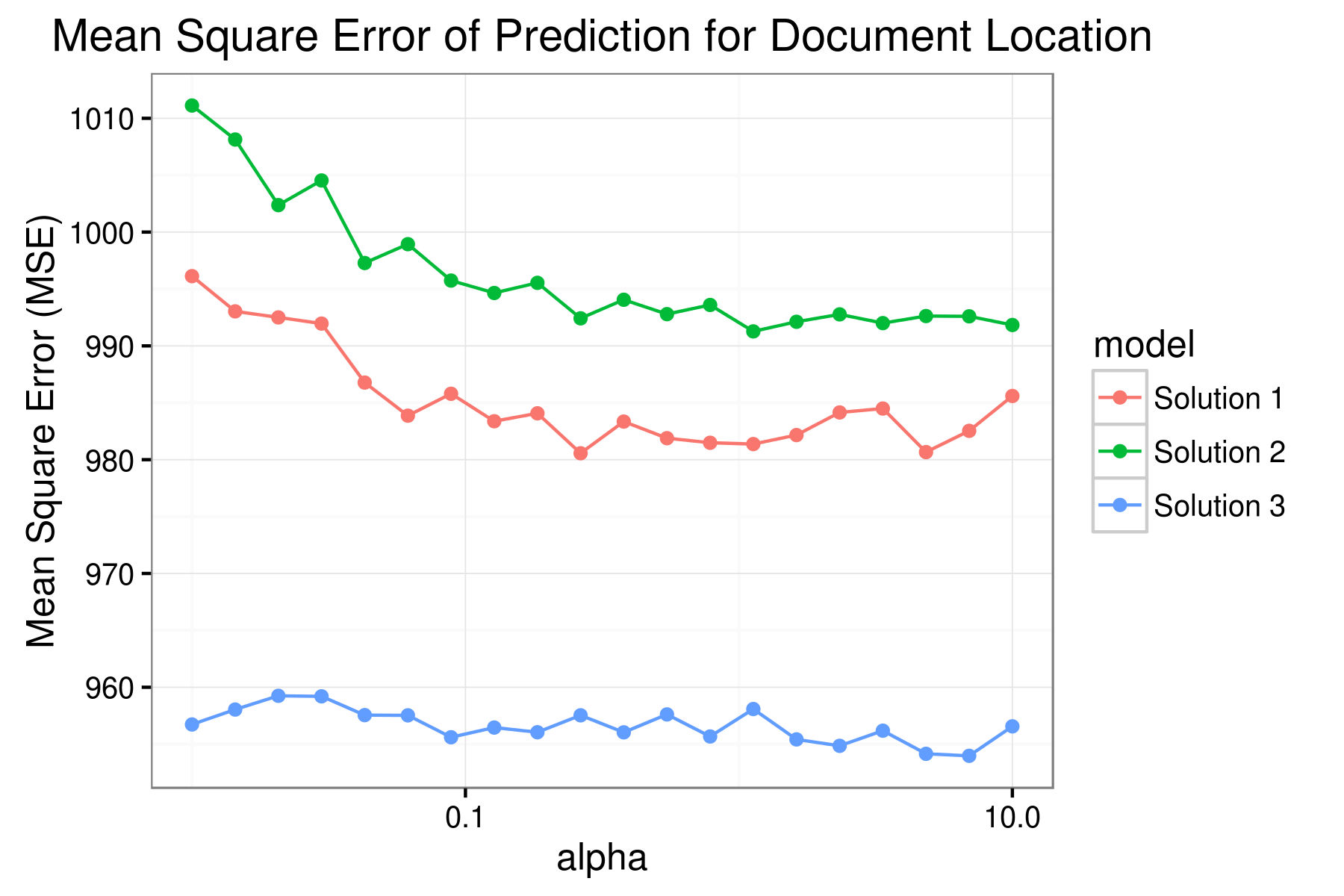}\par 
  \end{subfigure}

  \begin{subfigure}[b]{0.24\textwidth}
    \includegraphics[width=\columnwidth]{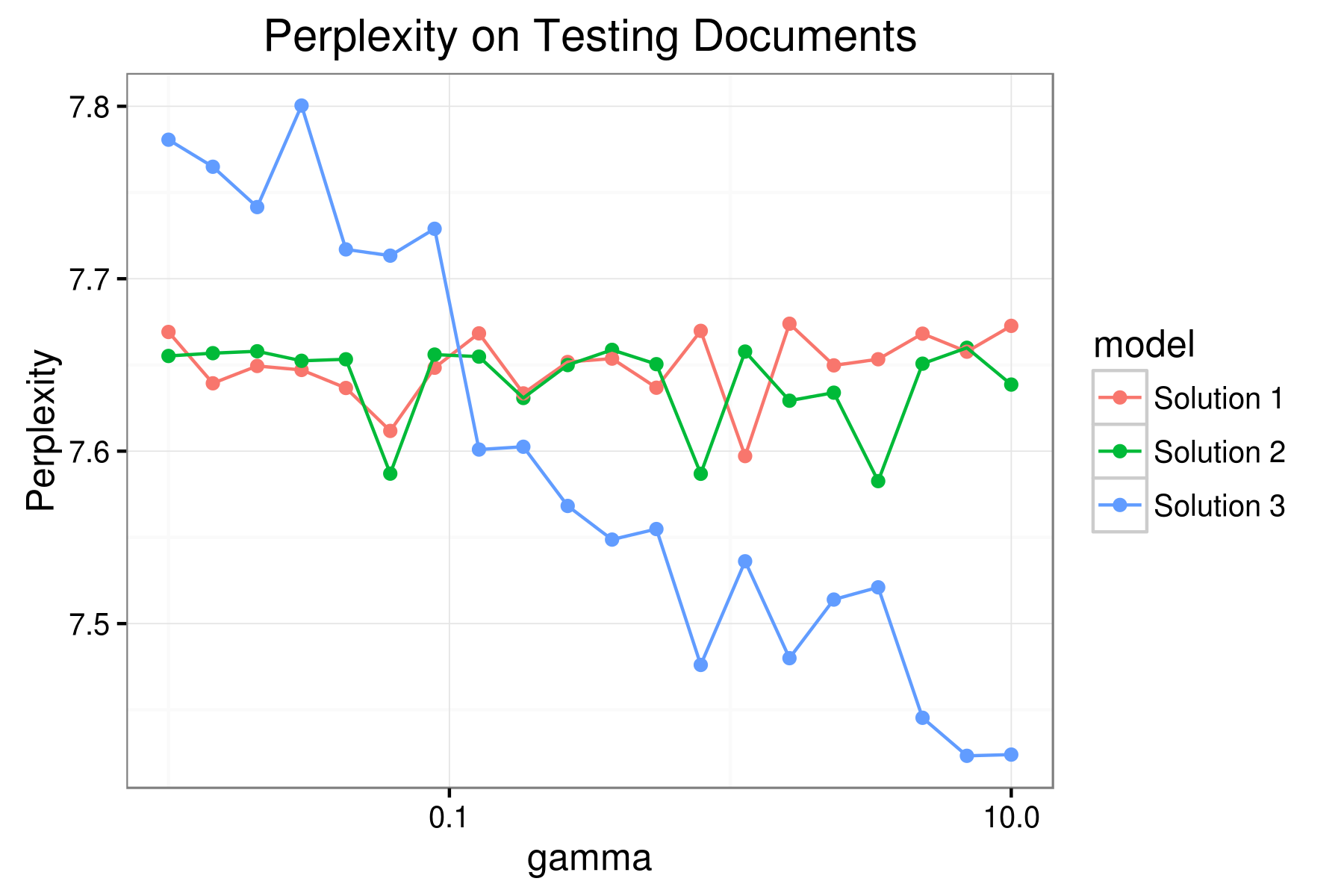}\par 
    \includegraphics[width=\columnwidth]{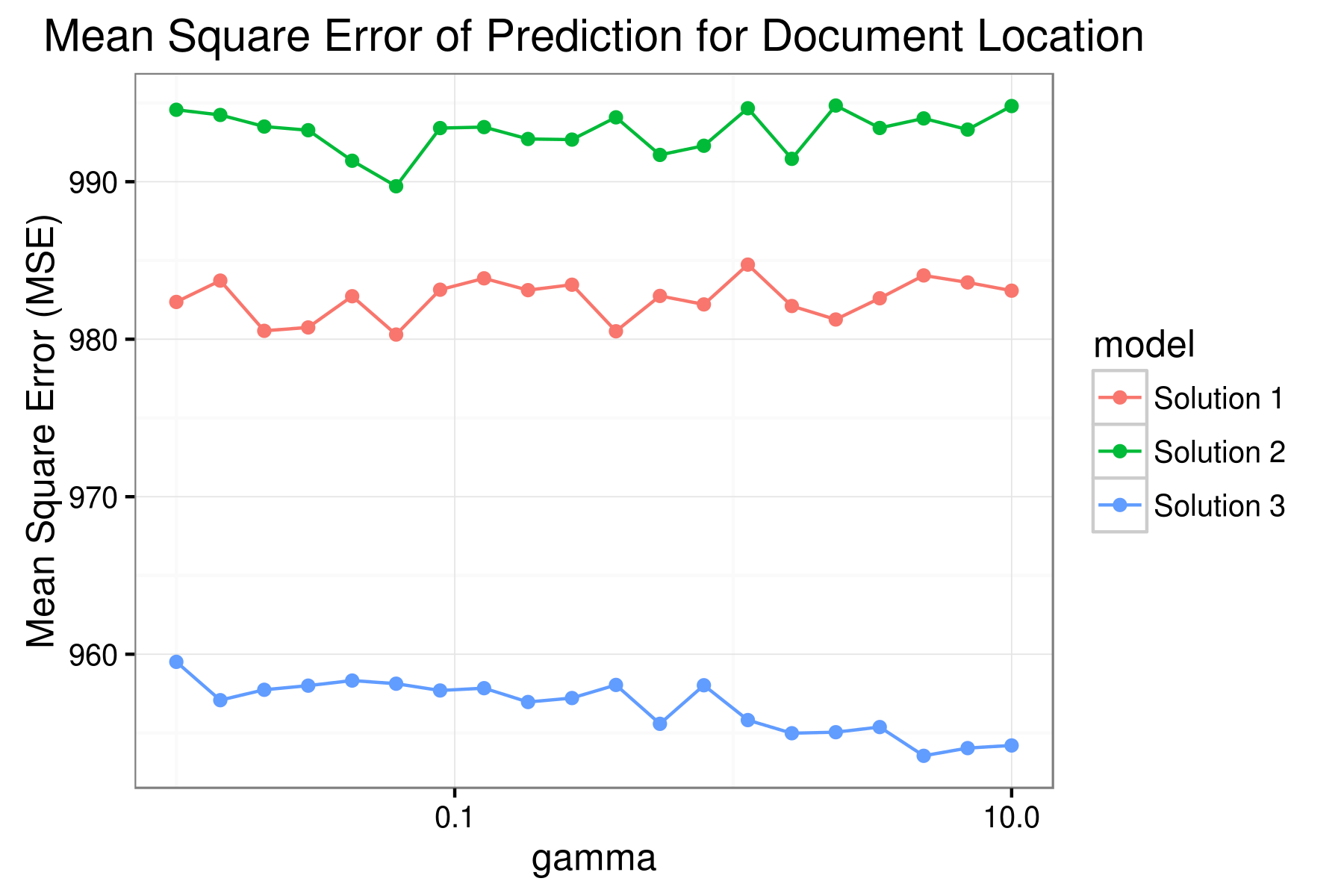}\par 
  \end{subfigure}

  \begin{subfigure}[b]{0.24\textwidth}
    \includegraphics[width=\columnwidth]{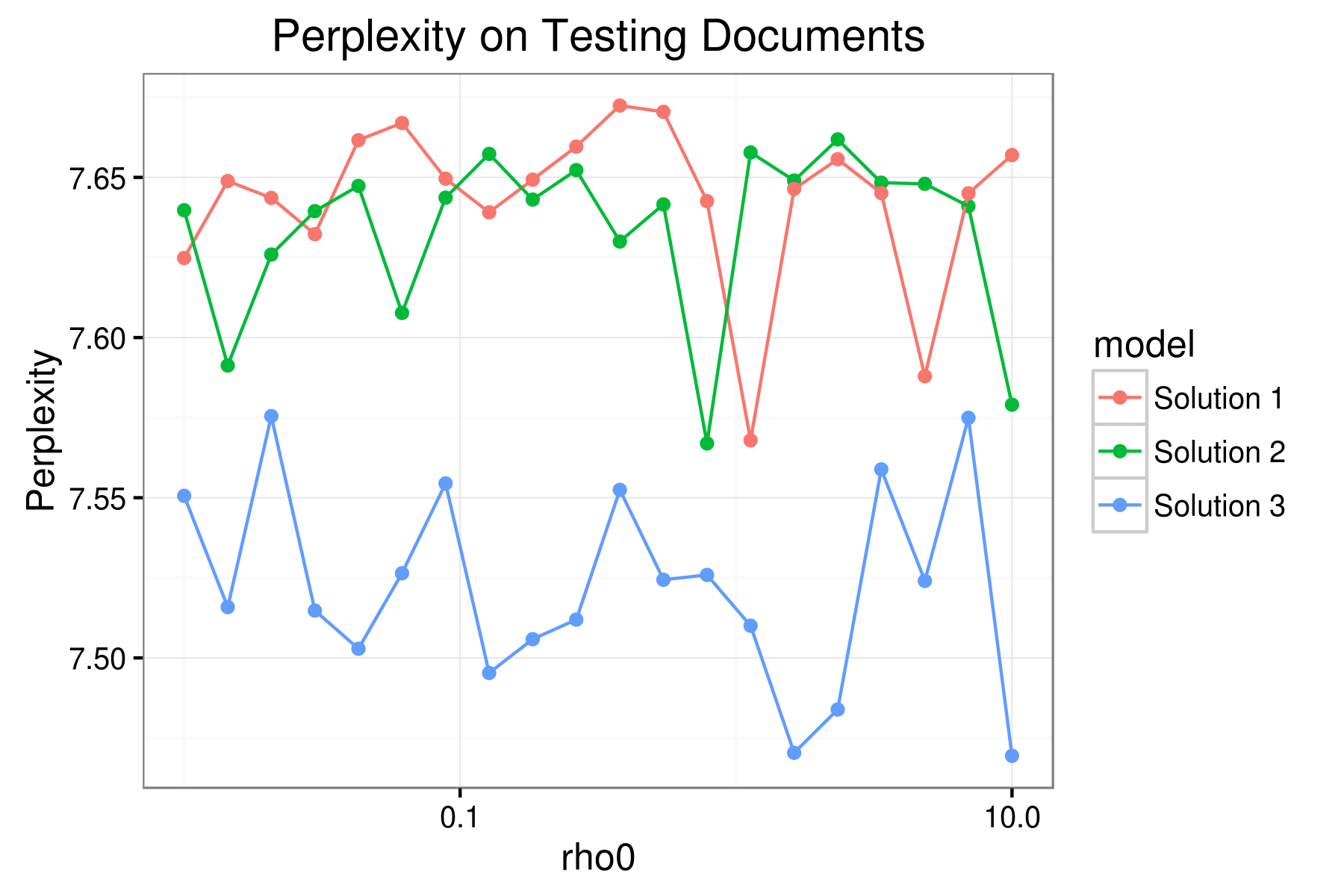}\par 
    \includegraphics[width=\columnwidth]{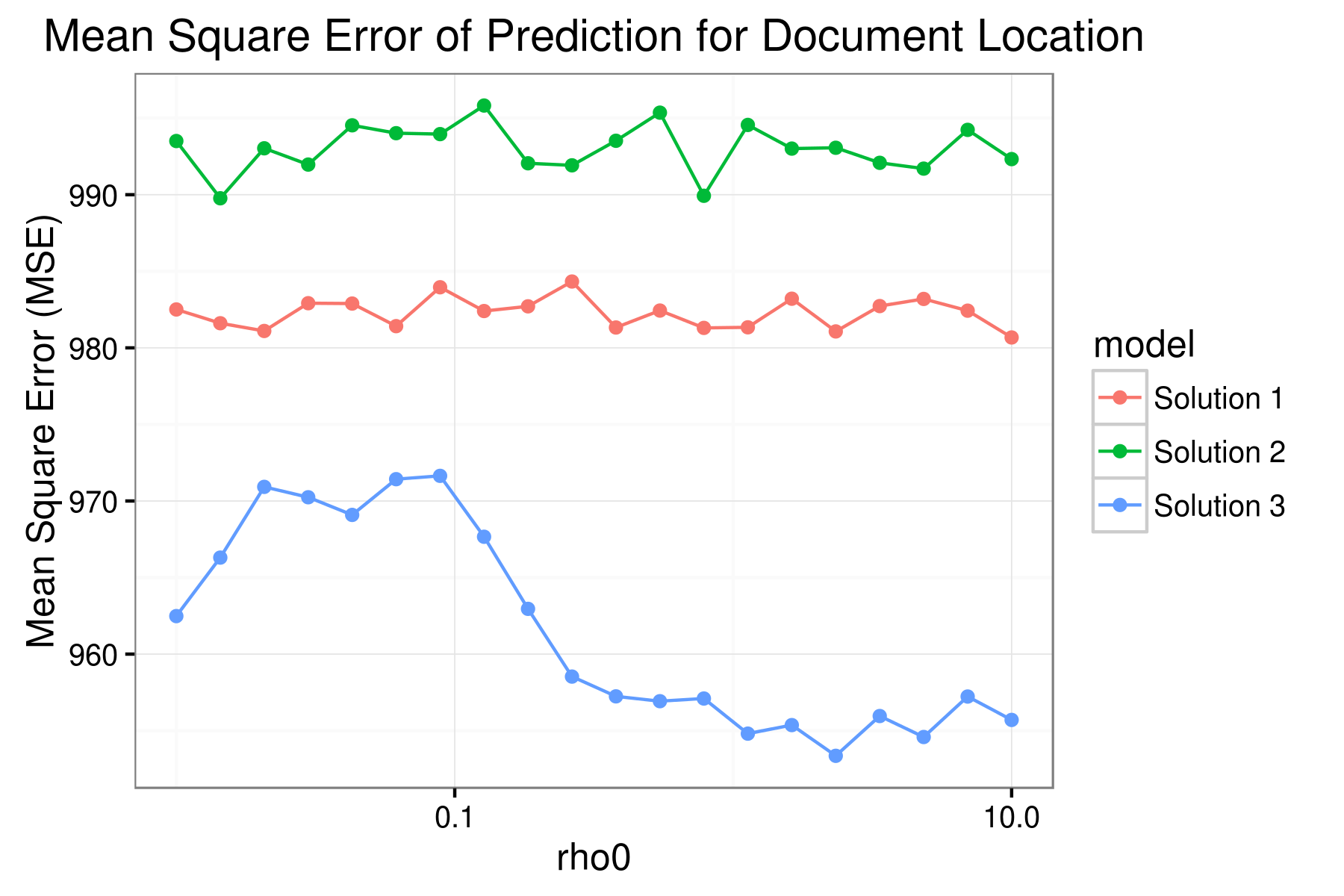}\par 
  \end{subfigure}

  \begin{subfigure}[b]{0.24\textwidth}
    \includegraphics[width=\columnwidth]{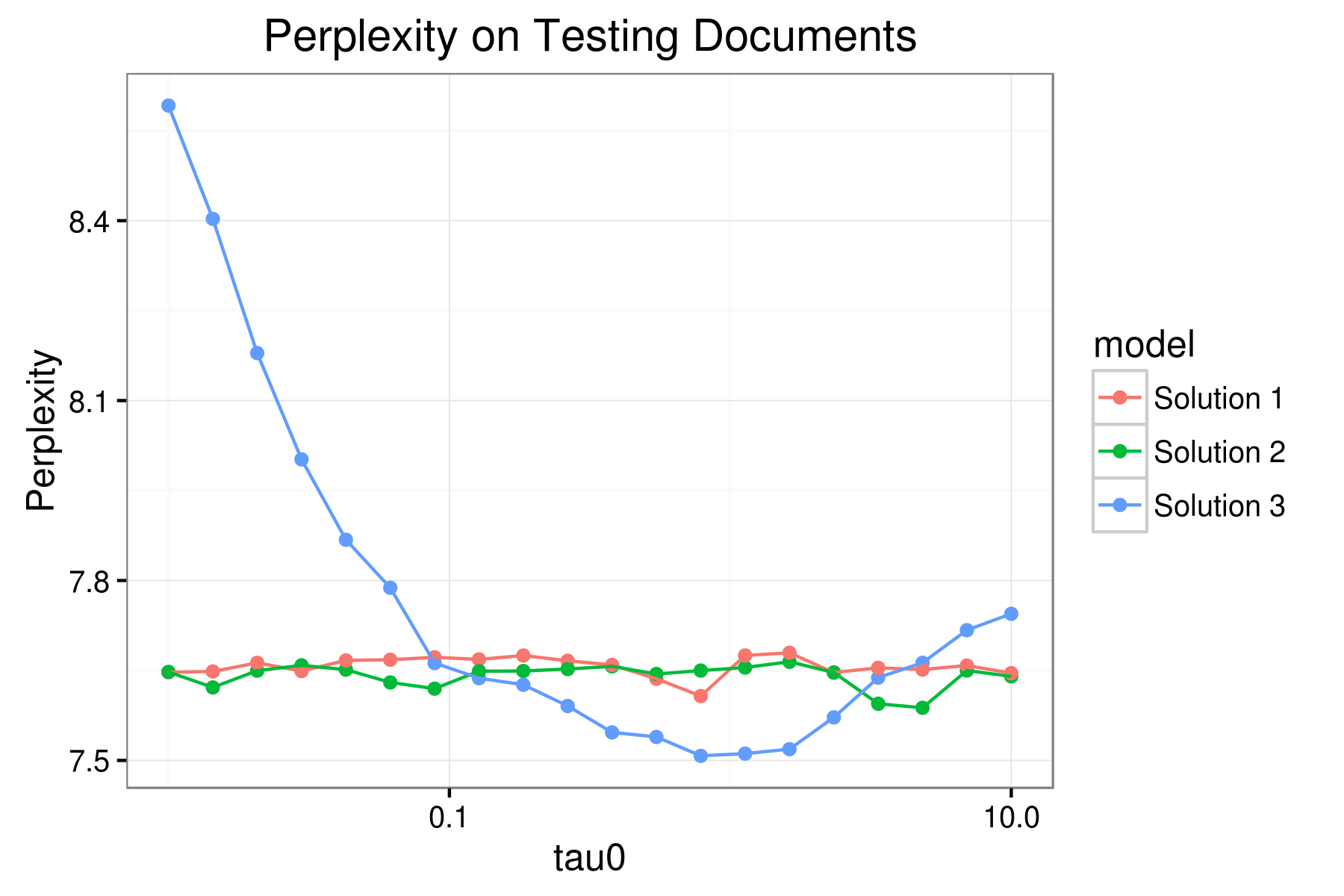}\par 
    \includegraphics[width=\columnwidth]{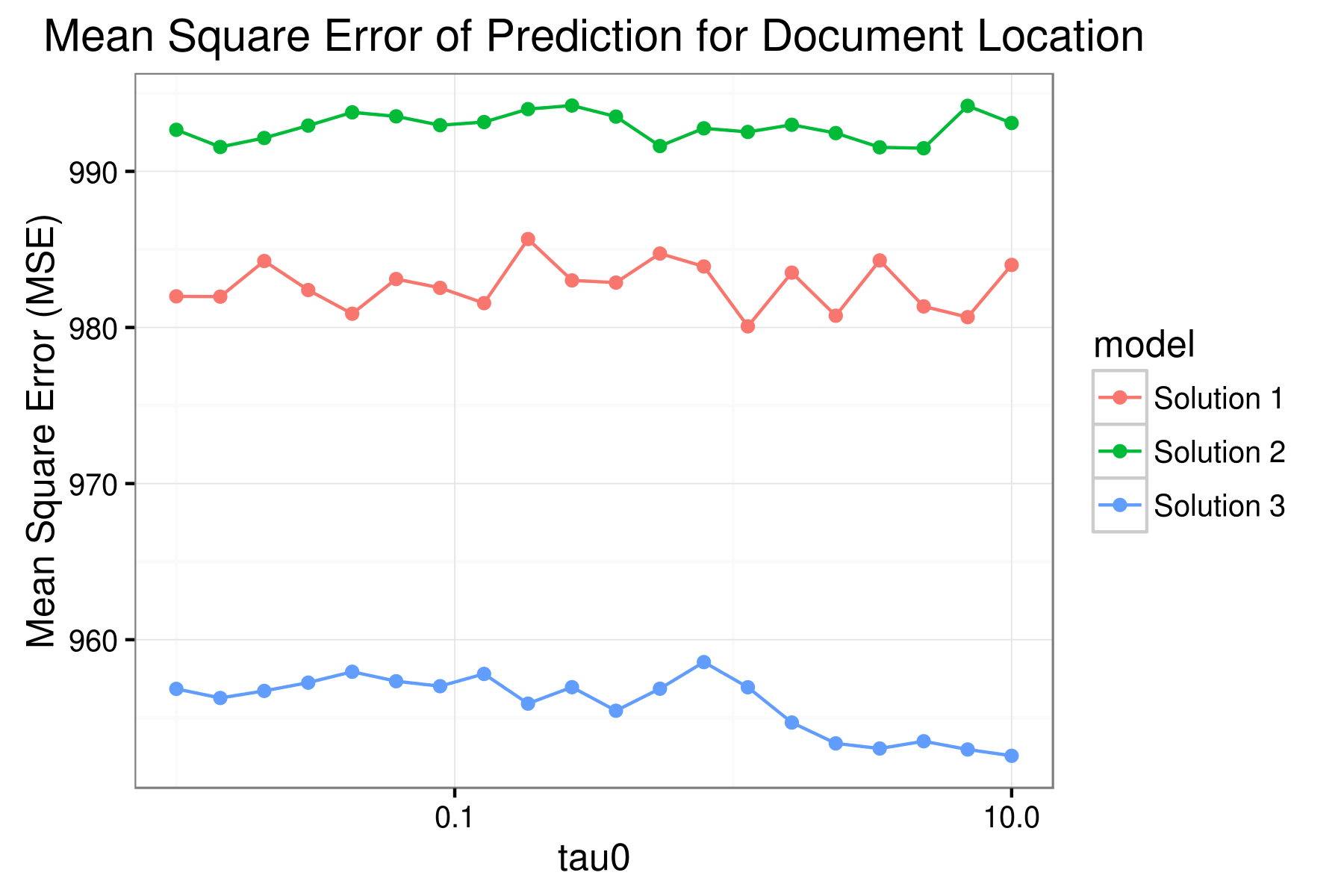}\par 
  \end{subfigure}

\end{multicols}
\caption{Perplexity and prediction for document location when model parameters are changed}
\label{fig2:prediction_results}
\end{figure*}

In the perplexity result in Figure~\ref{fig2:prediction_results}, we tested for results using various parameterizations of the model and the three different solutions for the Laplace approximation. Here, perplexity is calculated according to Equation~\ref{eq2:perp_def}. The first thing we see here is that \textbf{Solution 3} generally performs better than the other two baseline approaches. However, it is clear that this result is impacted by the setting of model parameters.

For example, perplexity on testing data changes with the parameter $\alpha$, which is the decay factor of the RCRP prior defined in Equation~\ref{eq2:def_s}- the perplexity when $\alpha$ is large is generally better than those with smaller $\alpha$. Our proposed approach (Solution 3) beats the other two baseline methods when $\alpha$ is larger than $0.1$. This indicates that our model prefers a less radical weight distribution on previous epochs. Instead, taking more epochs into considerations generates a much better result. 

We observe a similar pattern for variable $\gamma$, which is a dispersion parameter that controls the way that new clusters are created in the model. We see that a high $\gamma$ yields a better performance for our proposed approach, which is not surprising since generally the higher model complexity the better expressiveness the model will be able to generalize our data. 

The pattern of perplexity for variable $\tau_0$ is also very interesting. Recall that $\tau_0$ is the Markov transiting prior that controls the variance of the same topic from one time to the other. Since both the baseline solutions ignore this prior information, one of their performance will change with $\tau_0$. Our approach, however, does change with $\tau_0$ and we can clearly observe a region where our approach outperform the others. When $\tau_0$ is close to infinity, our proposed approach approximates \textbf{Solution 1}. When $\tau_0$ is close to zero, no counting on the current time step (i.e. $n_{t,k,i}$) is being used and nothing is being learned in the model. The Markov transiting prior for location, $\rho_0$ has less impacts to the perplexity results than the prediction error for location, which we will discuss in the next paragraph. 

\begin{equation}  
\begin{aligned} \label{eq2:perp_def}
perp (\mathcal{D}_{test})= - \frac{ \Sigma_t \Sigma_d \Sigma_{n} \log{p(w_{t,d,n} |s_{t,d}=k, \phi_{t,k})}}{ \Sigma_t \Sigma_d \Sigma_{n} 1 }
\end{aligned}
\end{equation}

Figure~\ref{fig2:prediction_results} shows the Mean Square Error (MSE) for the task of prediction the location specified of the left out posts. The results are illustrated in Figure~\ref{fig2:prediction_results}. Here we see that our proposed method (Solution 3) outperform the baselines significantly throughout the range of the tested parameters. Most parameters do not impact location prediction results. The only exception is $\rho_0$, which is the Markov transitioning prior for location. Here we see that the MSE climbed slightly with an increase of $\rho_0$ but declined sharply after $\rho_0$ reaches $0.1$. This indicates that higher prior values that put more weight on recent counting information are beneficial for the model to effectively learn to predict locations. 

%
%

\section{Conclusion}
In this paper, we proposed a Bayesian non-parametric model to discover evolutionary social events. We experimented with the model on Twitter data and are able to identified evolutionary latent social events that change over time - an example of the Super Bowl was given to highlight our model's ability to do so. There are several limitations to this paper that are open to potential future work. First, the assumption that each document has to belong to a specific event is a bit simplistic considering many tweets are not event-centric. Second, the fact that the spatial priors are predetermined makes the model difficult to deal with streaming data outside the predetermined spatial regions. And finally, spatial components do not penalize towards their distances to the event centers and this assumption can sometime generate universal events that doesn't contain any specific event information. Future word should address these issues.

\bibliographystyle{ACM-Reference-Format}
\bibliography{sigproc} 

\end{document}